\documentclass{article}
\usepackage[numbers,compress]{natbib}
\usepackage[final]{neurips_2025}

\usepackage[utf8]{inputenc} % allow utf-8 input
\usepackage[T1]{fontenc}    % use 8-bit T1 fonts
\usepackage{hyperref}       % hyperlinks
\usepackage{url}            % simple URL typesetting
\usepackage{booktabs}       % professional-quality tables
\usepackage{amsfonts}       % blackboard math symbols
\usepackage{nicefrac}       % compact symbols for 1/2, etc.
\usepackage{microtype}      % microtypography
\usepackage{xcolor}         % colors

\usepackage{graphicx}
\usepackage{multirow}
\usepackage{amsmath}
\definecolor{cvprblue}{rgb}{0.21,0.49,0.74}
\usepackage{footnote}
\makesavenoteenv{tabular}
\makesavenoteenv{table}
\usepackage{bm}

\definecolor{mgreen}{HTML}{D5E8D4}
\definecolor{mred}{HTML}{F8CECC}
\definecolor{airforceblue}{HTML}{5D8AA8}

\title{Controllable Human-centric Keyframe Interpolation with Generative Prior}

\author{%
    Zujin Guo$^{1}$\hspace{0.1in} 
    Size Wu$^{1}$\hspace{0.1in} 
    Zhongang Cai$^{2}$\hspace{0.1in} 
    Wei Li$^{1\ast}$\hspace{0.1in} 
    Chen Change Loy$^{1\ast}$\hspace{0.1in}  \\
    $^1$S-Lab, Nanyang Technological University\\
    $^2$SenseTime Research\\
    {\tt\url{https://gseancdat.github.io/projects/PoseFuse3D_KI}}
}

\begin{document}

\maketitle
\let\thefootnote\relax\footnotetext{$^{\ast}$ Corresponding authors}

\begin{abstract}
Existing interpolation methods use pre‑trained video diffusion priors to generate intermediate frames between sparsely sampled keyframes. In the absence of 3D geometric guidance, these methods struggle to produce plausible results for complex, articulated human motions and offer limited control over the synthesized dynamics.
In this paper, we introduce PoseFuse3D Keyframe Interpolator (PoseFuse3D-KI), a novel framework that integrates 3D human guidance signals into the diffusion process for Controllable Human-centric Keyframe Interpolation (CHKI). 
To provide rich spatial and structural cues for interpolation, our PoseFuse3D, a 3D‑informed control model, features a novel SMPL‑X encoder that transforms 3D geometry and shape into the 2D latent conditioning space, alongside a fusion network that integrates these 3D cues with 2D pose embeddings.
For evaluation, we build CHKI-Video, a new dataset annotated with both 2D poses and 3D SMPL‑X parameters. We show that PoseFuse3D-KI consistently outperforms state-of-the-art baselines on CHKI-Video, achieving a 9\% improvement in PSNR and a 38\% reduction in LPIPS.
Comprehensive ablations demonstrate that our PoseFuse3D model improves interpolation fidelity. 

\end{abstract}
\section{Introduction}
\label{sec:intro}

Frame interpolation aims to generate new frames between two consecutive video frames to improve temporal smoothness.
Traditional interpolation methods~\cite{guo2025generalizable,xue2019vimeo90k,jiang2018superslomo} assume small, simple motion over short time spans. These methods are challenged when the input frames are widely separated, known as keyframe interpolation or generative inbetweening~\cite{zhu2024generative,yang2024vibidsampler,wang2024generative,feng2024trf}, where the motion between them becomes complex and ambiguous. This challenge is magnified in human‑centric videos, where articulated body movements encompass diverse poses and shapes. With human subjects being prevalent in today’s video content, there is a growing need for interpolation methods to handle large temporal gaps and intricate human motion while offering plausible results.

\begin{figure*}[!t]
  \centering
  \includegraphics[width=1.0\linewidth]{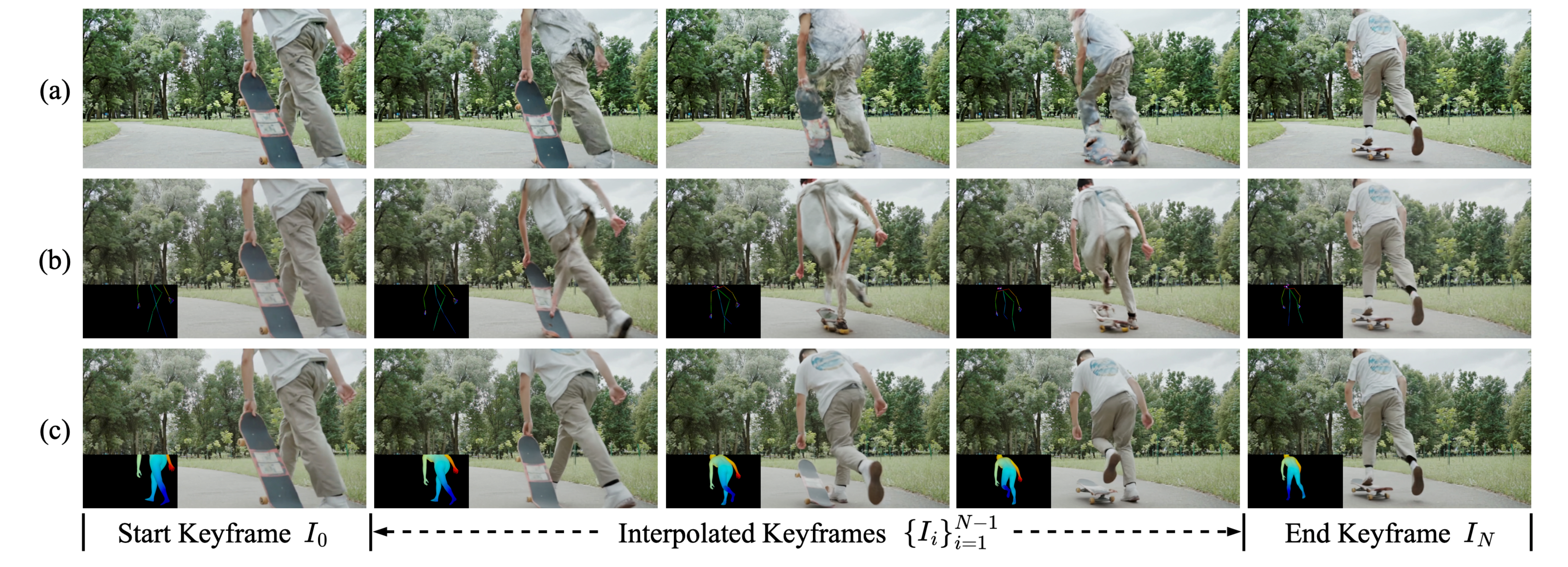}
  \vspace{-20pt}
  \caption{\textbf{Keyframe Interpolation with Different Strategies.} 
  (a) Interpolation using I2V models without intermediate guidance often yields implausible or distorted frames, especially under large motion or occlusion.
(b) Skeleton-guided interpolation offers structural cues but lacks geometric detail, resulting in unrealistic body shape and appearance.
(c) \textbf{Our PoseFuse3D-KI} employs dense human-centric guidance, enabling temporally coherent and visually plausible interpolations. 
}
  \vspace{-5pt}
  \label{fig:teaser}
\end{figure*}

Human‑centric keyframe interpolation remains challenging for current methods. Recent approaches~\cite{feng2024trf, yang2024vibidsampler, wang2024generative} leverage generative priors from image‑to‑video (I2V) models to bridge the temporal gap. These methods condition the interpolation solely on the input keyframes without intermediate guidance. Consequently, they often struggle to resolve motion ambiguities and accurately capture the complex articulated dynamics of human motion. For instance, when keyframes involve large occlusions or non‑rigid joint movements, these methods often produce implausible or distorted interpolations (Figure~\ref{fig:teaser}(a)) due to insufficient intermediate guidance. 
FCVG~\cite{zhu2024generative} has explored interpolation keyframes with 2D skeletons as control signals for human subjects. However, 2D lines cannot convey full body shape and geometry, leading to unrealistic results (see Figure~\ref{fig:teaser}(b)). 
These methods lack fine-grained control over the interpolation process, limiting their ability to produce flexible, high‑fidelity human‑centric interpolations.

In this study, we investigate the integration of 3D human conditions into the human‑centric keyframe interpolation pipeline. Drawing inspiration from recent advances in human animation~\cite{zhu2024champ, zhou2024realisdance}, we propose to integrate 2D human poses~\cite{yang2023dwpose} with 3D SMPL-X models~\cite{pavlakos2019smplx} as intermediate control signals. 
These signals provide precise guidance for complex articulated motions: 2D poses offer concise representations of human joint poses, while 3D models capture rich spatial geometry.
However, effectively processing these control signals poses challenges. First, common practice renders 3D human models into 2D proxies (e.g., colored surface, normals, depth maps) before encoding, leading to substantial loss of spatial information in occluded regions. Therefore, we need to develop a dedicated encoder that preserves occluded 3D details when converting models into control signals.  Second, fusing signals with different information content and granularity is nontrivial.
This necessitates designing appropriate neural architectures that can accurately extract 3D cues and harmonize them with 2D poses into a unified, informative control input.

To this end, we introduce PoseFuse3D Keyframe Interpolator, termed as \emph{PoseFuse3D-KI}, a novel pose-control framework for controllable human-centric keyframe interpolation (Figure~\ref{fig:teaser}(c)). Our framework is unique in its 3D-informed control model PoseFuse3D, which comprises three jointly trained modules. The first derives control features from visualized conditions; the second encodes and aggregates 3D body geometry into 2D image conditioning features; and the third combines outputs from the first two modules into a unified control signal for interpolation. 
In contrast to Champ~\cite{zhu2024champ}, which fuses 3D features from rendered visualizations, our encoder processes features directly in 3D and integrates projected features through feature aggregation.

To evaluate the proposed PoseFuse3D-KI, we have also built a high-quality video dataset for Controllable Human-centric Keyframe Interpolation (CHKI). Most existing interpolation datasets, such as SportsSlomo~\cite{chen2024sportsslomo}, target small temporal gaps, lack annotations for 2D poses or 3D human models, and offer limited human‑centric motion diversity. Therefore, we introduce \emph{CHKI-Video}, a new dataset for a systematic evaluation of CHKI algorithms. CHKI-Video comprises 2,614 high-quality
video clips of over 180K frames sourced from SportsSlomo~\cite{chen2024sportsslomo} and Pexels~\cite{pexels} website that hosts high-quality stock videos. Each frame is carefully annotated with bounding boxes, segmentation masks, 2D human poses, and SMPL‑X parameters, using state‑of‑the‑art tools~\cite{cai2023smpler,khirodkar2024sapiens,liu2024groundingdino,ravi2024sam2,wu2023dover} supplemented by manual verification.
From this collection, we derive a benchmark specifically for the CHKI task. We hope this benchmark will help improve
controlled keyframe interpolation techniques with its high-quality videos and diverse examples.

Our contributions are threefold:
(i) We present PoseFuse3D-KI, an effective interpolation framework for human-centric keyframe interpolation, characterized by a novel pose-control model, PoseFuse3D. It effectively extracts control signals from 3D SMPL-X and fuses 2D signals, allowing precise and informative control. 
(ii) For evaluation, we construct CHKI‑Video, a benchmark dataset with comprehensive human‑centric annotations, which are absent in existing interpolation benchmarks.
(iii) Through extensive experiments, we demonstrate that PoseFuse3D-KI delivers state‑of‑the‑art performance on our CHKI-Video benchmark with an improvement of 1.85 dB in PSNR and a reduction of 0.0796 in LPIPS.
\section{Related Work}
\label{sec:relwork}

\noindent\textbf{Frame Interpolation.} 
Traditional frame interpolation methods are primarily designed for temporally adjacent frames and rely on either direct synthesis using convolutional networks~\cite{choi2020snufilm, kalluri2023flavr}, or motion representations such as dynamic kernels~\cite{Niklaus_CVPR_2017, Niklaus_ICCV_2017, peleg2019net, EDSC, ding2021cdfi, lee2020adacof} and optical flows~\cite{huang2022rife, li2023amt, Kong_2022_ifrnet, siyao2021deep, hou2023video, DAIN, guo2025generalizable, jiang2018superslomo, xu2019quadratic, zhang2023emavfi}. 
Recent advances~\cite{feng2024trf,yang2024vibidsampler,wang2024generative} extend this task to more challenging keyframe interpolation scenarios by leveraging the generative priors of image-to-video diffusion models~\cite{blattmann2023svd,wang2025wan,rombach2022high,ho2020denoising}.
These methods combine temporal forward and reverse denoising predictions in a unified process to enable interpolation.
However, such approaches struggle when faced with complex articulated motions or ambiguous transitions from human keyframes.
To alleviate motion ambiguity in interpolation, FCVG~\cite{zhu2024generative} introduces 2D matched lines as control signals. However, this control signal lacks the 3D geometric context required for plausible human-centric interpolation, resulting in unrealistic body shape and appearance.

\noindent\textbf{Pose-Guided Human Animation.}
Recent advances in pose-guided human animation~\cite{zhou2024realisdance,zhu2024champ,hu2024animateanyone,peng2024controlnext,zhang2023controlnet,xu2024magicanimate,wang2024disco} harness the power of diffusion models and have achieved remarkable success in generating videos from a single reference image. These methods offer flexible and precise control by incorporating enriched conditioning signals and increasingly sophisticated control mechanisms.
2D human poses, such as OpenPose~\cite{cao2017openpose} and DWPose~\cite{yang2023dwpose}, are widely used in existing works~\cite{zhu2024champ, zhou2024realisdance, peng2024controlnext, zhang2023controlnet}, but they are limited in capturing fine-grained geometry and motion dynamics. To address this, recent works~\cite{zhu2024champ, zhou2024realisdance} integrate 3D human parametric models~\cite{loper2023smpl, pavlakos2019smplx}, which offer realistic body representation through blend shapes and skinning, resulting in more accurate and expressive human animations.
Specifically, Champ~\cite{zhu2024champ} renders 3D human models into 2D proxies (e.g., normals, depth, and semantic maps) and combines them with 2D pose visualizations as control input. It operates directly on these visualizations and unifies them using a simple summation operation.
To incorporate control signals into diffusion models, many approaches~\cite{wang2024disco,xu2024magicanimate,zhang2023controlnet,zhou2024realisdance} adapt ControlNet~\cite{zhang2023controlnet} for their customized control networks. 
Some methods~\cite{zhu2024champ,hu2024animateanyone} introduce task-specific pose guiders but often require retraining the majority of denoiser parameters. 
ControlNeXt~\cite{peng2024controlnext} improves control efficiency by encoding conditions with a lightweight convolutional network and injecting them via cross-normalization after the first denoising block, tuning only a minimal subset of parameters. This efficient mechanism enables robust conditioning over large-scale pre-trained video generators~\cite{wang2025wan,blattmann2023svd}. 

In this paper, we demonstrate the advantage of combining 2D poses (DWPose~\cite{yang2023dwpose}) with 3D human models (SMPL-X~\cite{pavlakos2019smplx}) through comprehensive experiments. We present a novel pose control model that extracts unified, 3D-informed control features to provide precise guidance. Instead of relying on rendered normals and depth maps, the model directly extracts explicit 3D information from human models in 3D space, preserving richer control signals. It adopts a ControlNeXt-inspired strategy to control video diffusion models for keyframe interpolation.

\section{Method}

Given a human-centric keyframe-pair $I_0, I_N\in \mathbb{R}^{H\times W \times3}$ with timesteps $\{0,N\}$, Controllable Human-centric Keyframe Interpolation (CHKI) is formulated as:
\begin{align}
  & \{\hat{I}_i\}^{N-1}_{i=1} = \mathcal{G}(I_0,I_N,\{\textbf{C}\}^{N}_{i=0}),
\label{eq:vfi}
\end{align}
where $\mathcal{G}$ denotes an interpolation model guided by control signals $\{\textbf{C}_i\}$.
In this work, we aim to address CHKI by proposing an effective controllable interpolation framework, \textbf{PoseFuse3D-KI}. This framework integrates 3D-aware human-centric signals into a pre-trained Video Diffusion Model (VDM) through our 3D-informed control model PoseFuse3D, as illustrated in Figure~\ref{fig:main}(a).

\subsection{PoseFuse3D}
PoseFuse3D is a 3D-informed control model that provides 3D human structure and geometry guidance for plausible human interpolation. 
This 3D-informed guidance is injected into the base diffusion model after the first denoising block via cross-normalization~\cite{peng2024controlnext}. 
Internally, PoseFuse3D comprises three jointly trained components: a visual encoding module that derives control features from 3D SMPL-X~\cite{pavlakos2019smplx} renderings and 2D DWPose~\cite{yang2023dwpose} visualizations; a SMPL‑X encoder that directly embeds 3D humans and aggregates them into image conditioning maps; and a fusion module that integrates encoding streams into a unified control tensor to guide interpolation in the base VDM.

\begin{figure*}[!t]
  \centering
  \includegraphics[width=0.95\linewidth]{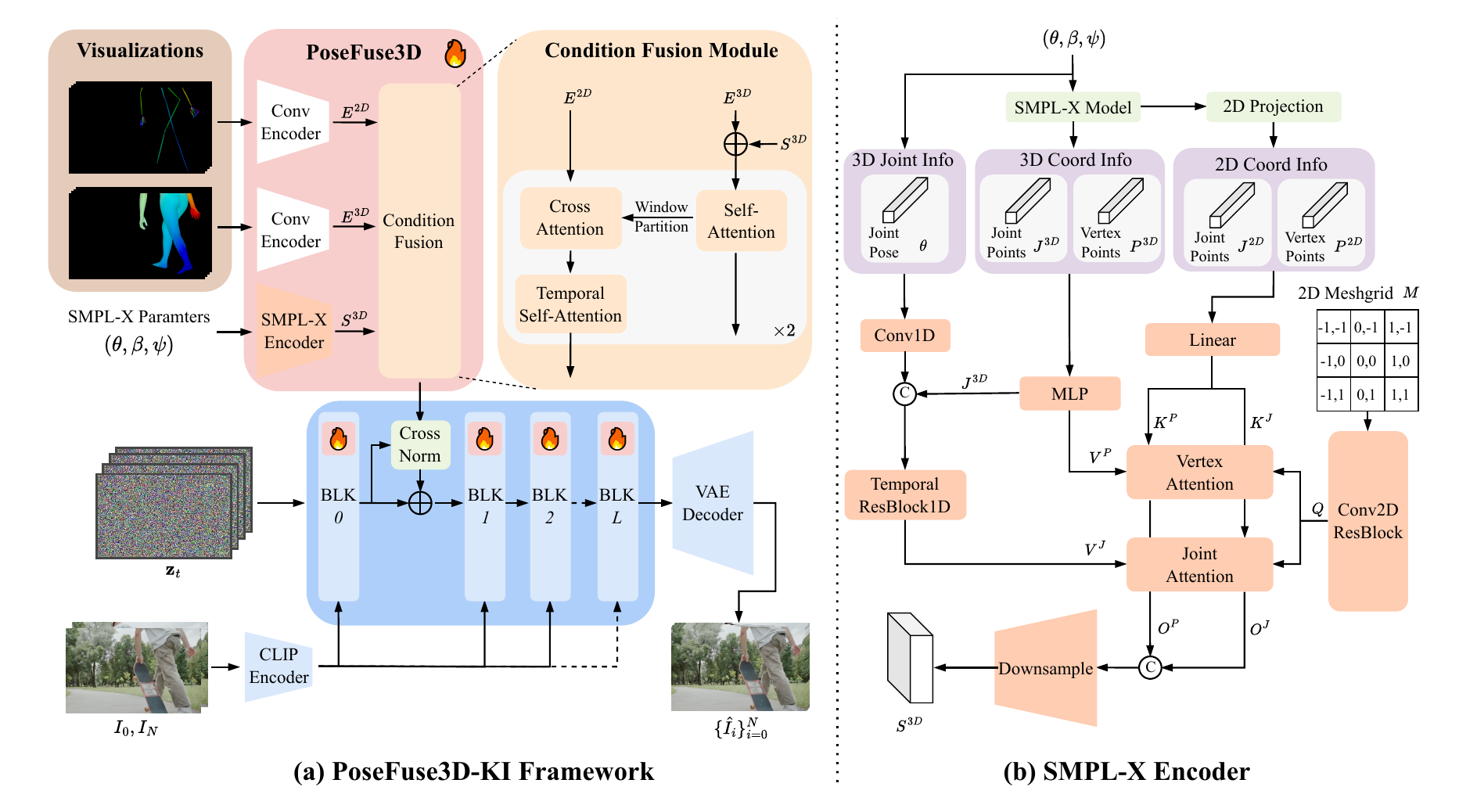}
  \vspace{-10pt}
  \caption{\textbf{Model Architecture.
  }
  Our PoseFuse3D-KI framework, as shown in (a), comprises a video diffusion model (VDM) and a novel control model, PoseFuse3D. The PoseFuse3D model extracts rich features from both 3D and 2D control signals and fuses them into a unified representation to guide the VDM. The key component of PoseFuse3D is the SMPL-X encoder as illustrated in (b), which provides \textit{explicit 3D signal features}. Specifically, the SMPL-X encoder first extracts 3D information from the SMPL-X model with 2D correspondences via projection. The 3D and 2D information is then encoded in parallel. With features of 2D correspondences, 3D information is aggregated onto the 2D image plane using attention mechanisms. The aggregated features are subsequently processed to produce the final feature $S^{3D}$.
  }
  \label{fig:main}
  \vspace{-10pt}
\end{figure*}

\noindent\textbf{Visual Encoding.} 
This visual encoding module extracts conditional features from visualized control images, maintaining natural pixel-level alignment with the controlled video latents. To enhance the control signals, we incorporate visualizations from both DWPose and SMPL-X. While SMPL-X renderings provide rich human surface details, their keypoint information is indirect, mixed with other vertices and mesh faces. Therefore, we add DWPose visualizations to emphasize the skeletal keypoint layout for robust pose understanding. This combination was also demonstrated to be effective in prior work~\cite{zhu2024champ}. 
Specifically, our visual encoding module employs two parallel convolutional encoders to capture comprehensive pose information. 
One encoder processes DWPose visualizations to capture compact pose information, while the other handles SMPL-X renderings that retain 3D cues such as occlusion boundaries and projected shapes. 
Notably, to enrich semantic detail, we use SMPL-X Colored Surface~\cite{zhou2024realisdance} during rendering, which assigns a unique color to each vertex. The resulting feature maps, $E^{2D}$ and $E^{3D}$, are then passed to the fusion module for unified conditioning.

\noindent\textbf{SMPL-X Encoder.}
Although 3D model renderings offer aligned image maps for conditioning in VDMs, the rendering operation discards parts of 3D information, particularly in \textit{occluded regions}. This results in implausible interpolation for keyframes of large human motion.
For example, in the first case of Figure~\ref{fig:vis_3dctrl}, all variants relying solely on rendered controls fail to interpolate the occluded arms with correct spatial position.
To enhance controllability with direct 3D information, we introduce a SMPL‑X encoder (Figure~\ref{fig:main}(b)) that processes the SMPL‑X model in 3D space and transforms it into an image conditioning feature $S^{3D}$. Specifically, a SMPL-X model~\cite{pavlakos2019smplx} is parameterized with $(\theta,\beta,\psi)$, corresponding to pose, shape, and expression.  We obtain structural information by forwarding these parameters into the SMPL‑X model to generate vertex and joint coordinates $P^{3D}, J^{3D}$ in 3D space and obtain their corresponding 2D coordinates $P^{2D}, J^{2D}$ through projection:
\begin{align}
  & P^{3D}, J^{3D} = \mathrm{SMPLX}(\theta,\beta,\psi),\\
  & P^{2D}, J^{2D} = \mathrm{Projection}(P^{3D}, J^{3D}),
\label{eq:proj}
\end{align}
where the joint coordinates $J^{3D}, J^{2D}$ correspond to the pose parameter $\theta$ that indicates joint rotations. Notably, the projection step establishes a correspondence between the 3D space and the 2D image plane, making it possible to retain 3D spatial structure while producing image conditioning maps.
Next, the raw 3D coordinates are processed using an MLP to produce point-wise vertex features $V^{P}$ and joint features. These joint features are refined through a temporal residual block that fuses them with pose information into expressive joint-level representations $V^{J}$. 
To aggregate these 3D features into 2D image control maps for joints and vertices, we employ separate attention mechanisms:
\begin{align}
  & O^{J} = \mathrm{JointAttn}(Q,K^{J},V^{J}),\\
  & O^{P} = \mathrm{VertexAttn}(Q,K^{P},V^{P}),
\label{eq:map}
\end{align}
where $Q\in\mathbb{R}^{B\times HW\times d}$ is the flattened $d$-dimension feature of a standard 2D meshgrid extracted by a convolutional encoder.
Finally, the outputs $O^{J}$ and $O^{P}$ are concatenated and passed through a downsampling block to produce the final SMPL‑X control representation $S^{3D}$, which serves as an informative and compact image embedding of the underlying 3D human structure.

\noindent\textbf{Condition Fusion.} 
The condition fusion module combines control features from both 2D and 3D signals into a unified representation to guide keyframe interpolation. For robust feature representation, we introduce a coarse-to-fine fusion strategy that progressively integrates rich geometric information from the 3D features into the compact 2D pose features. Specifically, we adopt two attention-based fusion blocks to perform this integration, where each block contains three attention layers for progressive refinement.  The first layer is a self-attention module that processes the 3D features by operating on the sum of rendering encoding $E^{3D}$ and SMPL-X features $S^{3D}$. 
The second layer performs cross-attention, aligning the 3D features with the 2D encoding $E^{2D}$ through a spatially localized interaction scheme. Notably, we adapt the shifted window-partition strategy~\cite{liu2021swin,zhang2023emavfi} to restrict attention computation to adjacent regions, enhancing local alignment. The third layer applies temporal self-attention to capture temporal dynamic correlations in the fused representation.
We use the second fusion block output as the final control signal, which is injected into the base interpolation engine to provide fine-grained, structure-aware guidance during synthesis.

\subsection{VDMs as Base Interpolation Engine}
\label{sec:adaptevdm}
To supply generative priors for human‑centric keyframe interpolation, we adapt pre‑trained Video Diffusion Models (VDMs) under the latent diffusion framework~\cite{rombach2022high}.
VDMs perform the diffusion process in a VAE‑encoded latent space, conditioning video synthesis on input frames. 
Our main experiments investigate Wan2.1~\cite{wang2025wan} as the base model for keyframe interpolation.

\noindent\textbf{Wan2.1} consists of a scaled‑up DiT‑based denoiser~\cite{peebles2023dit} and a causal 3D-VAE that performs spatiotemporal compression. It employs the flow matching strategy~\cite{lipmanflowmatching, esser2024rectifiedflow} for the diffusion process. It formulates the forward diffusion process as a linear interpolation between the clean video latent $\textbf{z}$ and the noise $\bm\epsilon$, which adds the noise by: $\textbf{z}_t=(1-t)\textbf{z}+t\bm\epsilon$. In the backward process, the denoiser $f_\theta$ iteratively refines $\mathbf{z}_t$ conditioned on the first frame $I_0$.
The training objective is formulated as:
\begin{align}
  & \mathcal{L} = \mathbb{E}_{\textbf{z},t,I_0,\bm{\epsilon}\sim\mathcal{N}(0,I)}[||f_\theta(\textbf{z}_t,I_0,t)-\textbf{y} ||_2^2],
\label{eq:obj}
\end{align}
where the target objective $\textbf{y}$ is $\frac{d\textbf{z}_t}{dt}=\bm\epsilon-\textbf{z}$. 
Since the latent space is unevenly compressed in time, noise fusion strategies~\cite{feng2024trf,zhu2024generative} that combine temporally forward and reverse denoising paths are ineffective.
Instead, we adapt Wan2.1 for keyframe interpolation in one unified denoising process. Specifically, we condition Wan2.1 on both endpoint frames $I_0, I_N$, and apply LoRA tuning to its input patch embeddings. Additional details are provided in the supplementary material.
\section{The CHKI-Video Dataset}
Existing interpolation datasets~\cite{xue2019vimeo90k, choi2020snufilm, chen2024sportsslomo, sim2021xvfi, stergiou2024lavib}, which are designed for interpolating temporally adjacent frames, are not suitable for the CHKI task.
To address this gap, we built the CHKI‑Video dataset in three consecutive stages. More details can be found in the supplementary material.

\noindent\textbf{Stage 1: Dataset Collection.} 
To create a dataset with challenging and diverse human motion, we curate video clips from SportSlomo~\cite{chen2024sportsslomo} and Pexels~\cite{pexels}. While SportsSlomo provides challenging human-centric videos, its exclusive focus on sports limits the diversity of activities. To enhance diversity versus the reality~\cite{yu2023dataset}, we compile a list of keywords, spanning everyday activities to high-intensity actions. We use these keywords to retrieve videos from Pexels, complementing the sports videos. The curated videos are then downsampled to eliminate frame redundancy unnecessary for keyframe interpolation.

\noindent\textbf{Stage 2: Pre-annotation Processing.} We first perform general filtering for the low-quality videos according to brightness changes and assessed scores~\cite{wu2023dover}.
Then, we use Grounding-DINO~\cite{liu2024groundingdino} and SAM2~\cite{ravi2024sam2} to detect, segment, and track human instances in each video. We discard any video of more than three people or that is shorter than 20 consecutive frames to ensure a sufficient temporal span for keyframe interpolation.
After automated processing, we manually review and filter detections in complex sports scenarios.

\noindent\textbf{Stage 3: Human-centric Annotation.} Building on the accurate human detections obtained in Stage 2, we annotate each clip for precise human-centric information. First, we employ Sapiens~\cite{khirodkar2024sapiens} to extract 2D human keypoints and perform whole‑body detection to filter out clips with incomplete human figures. This ensures our dataset remains strictly human-centric. Finally, we apply SMPLer‑X~\cite{cai2023smpler}, leveraging its high re‑projection accuracy to fit detailed SMPL‑X models for human images and produce reliable 3D body parameters for each frame.

As a result, CHKI-Video comprises 2,614 video clips of over 180K frames, carefully annotated with human bounding boxes, masks, 2D keypoints and 3D parametric models. 
To prepare the train and test split, we follow the original division for SportsSlomo~\cite{chen2024sportsslomo} videos and distribute the Pexels videos according to their keyword frequencies to maintain balanced coverage of all motion categories.
\section{Experiments}
We present quantitative and qualitative results in Sec.~\ref{sec:3dctrl} to validate the effectiveness of our 3D control strategy in PoseFuse3D. We compare our interpolation performance against state-of-the-art methods in Sec.~\ref{sec:bm} and analyze the scalability across temporal gaps in Sec.~\ref{sec:temporal_gaps}.
We further assess the in-the-wild interpolation capability in Sec.~\ref{sec:itw_interp_results}, where ground-truth control signals are not available.
Finally, Sec.~\ref{sec:ablation} provides a detailed ablation study to justify our model design.

\noindent\textbf{Implementation Details.}
We fine-tune PoseFuse3D-KI on the CHKI-Video training split for 70k iterations. Specifically, we fine-tune our 3D-informed control model PoseFuse3D, and employ LoRA on the input patch embeddings, as well as the value and output projections of the VDM’s attention modules. 
During training, we randomly sample 25 consecutive frames from video clips and process them to a resolution of $ 512\times320$. To maximize human-centric content, we crop each clip around its largest annotated human bounding box.  Given the ratio from the target input size, we first crop the maximum scale of the image with the largest box as the center. The cropped image is then resized to match the target resolution.
More implementation details are provided in the supplementary material.

\noindent\textbf{Evaluation Protocols.}
Unless otherwise noted, all methods are evaluated on the CHKI-Video test set using \textbf{ground-truth annotations} for controllable interpolation.
With motion ambiguity limited by the ground truth controls, we adopt the standard interpolation metrics of PSNR and LPIPS computed on the whole images for evaluation. To quantify performance specifically on human regions, we further leverage the annotated human boxes and masks, yielding PSNR\textsubscript{bbox}, LPIPS\textsubscript{bbox}, PSNR\textsubscript{mask}, and LPIPS\textsubscript{mask} metrics. Notably, we apply binary dilation to expand and smooth the masks before computing the metrics, preventing potential artifacts along mask boundaries.

\subsection{3D Control Strategy in PoseFuse3D}
\label{sec:3dctrl}
\begin{table}[t]
\caption{\textbf{Comparisons of 3D Control Strategies.} This table presents quantitative results of different 3D control strategies. The PSNR and LPIPS metrics are calculated for the whole image, as well as for the human-centric parts via annotated boxes or masks. Best results are highlighted with \textbf{boldface}.}
\label{tab:main_table}
\centering
\setlength\tabcolsep{3pt}
\resizebox{\textwidth}{!}{
\begin{tabular}{lcccccccc}
\toprule
\multirow{2}{*}{Method} & \multirow{2}{*}{Backbone} & \multirow{2}{*}{3D Control Strategy}  & \multicolumn{6}{c}{Evaluation Metrics} \\ \cmidrule(l){4-9}
& & & PSNR$\uparrow$  & PSNR\textsubscript{bbox}$\uparrow$ & PSNR\textsubscript{mask}$\uparrow$  & LPIPS$\downarrow$  & LPIPS\textsubscript{bbox}$\downarrow$  & LPIPS\textsubscript{mask}$\downarrow$\\ \midrule

FCVG & SVD & N.A. & 20.42 & 18.05 & 16.91 & 0.2100  & 0.0899 & 0.0606 \\
PoseFuse3D & SVD & VE-SVD & 20.96 & 18.56 & 17.47 & 0.1975  & 0.0835 & 0.0553 \\
\cmidrule{1-9}
PoseFuse3D & Wan2.1-I2V & VE & 21.91 & 19.13 & 17.87 & 0.1400  & 0.0682 & 0.0484 \\
PoseFuse3D & Wan2.1-I2V & VE+DN & 22.07 & 19.12 & 17.80 & 0.1363  & 0.0667 & 0.0473 \\
PoseFuse3D & Wan2.1-I2V & VE+SE (Ours) & \textbf{22.14} & \textbf{19.30} & \textbf{18.01} & \textbf{0.1330}  & \textbf{0.0653} & \textbf{0.0464} \\
\bottomrule
\end{tabular}
}
\vspace{-3pt}
\end{table}
\begin{figure*}[t]
  \centering
  \includegraphics[width=1.0\linewidth]{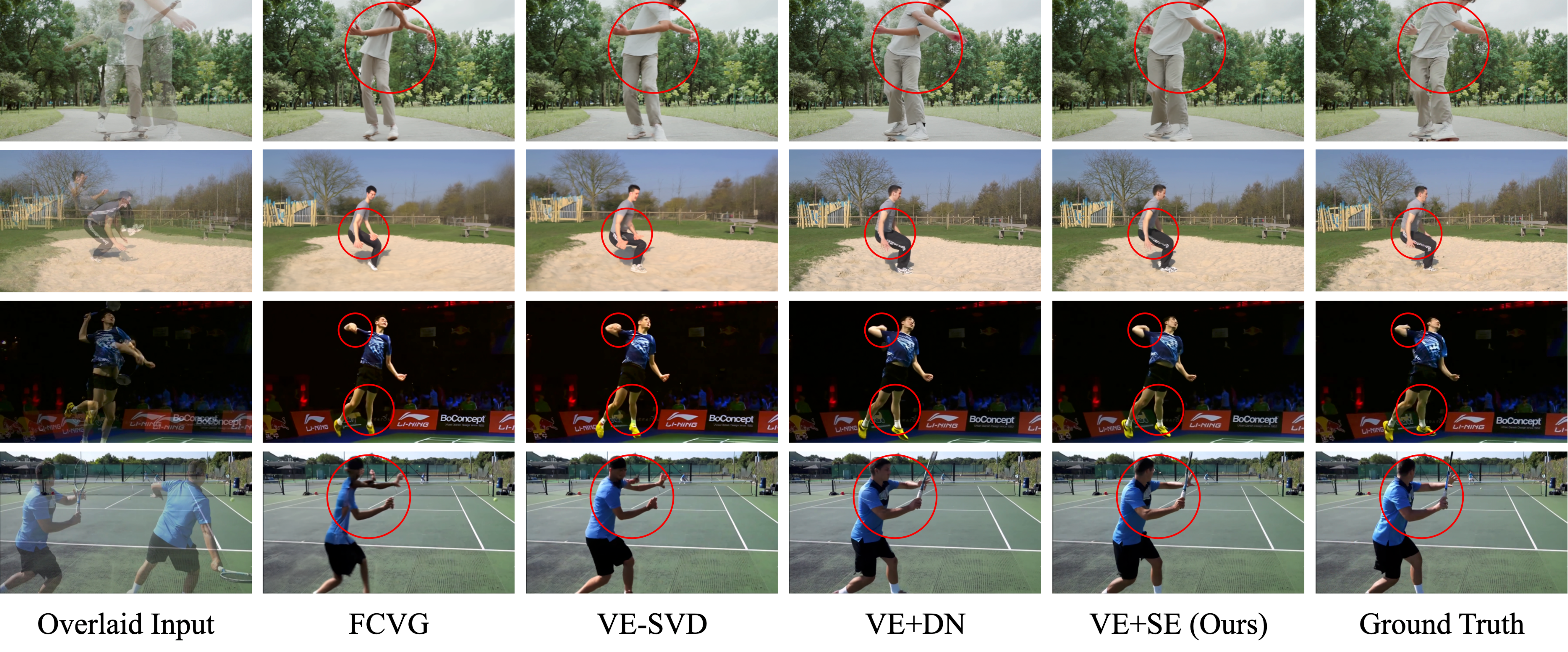}
  \caption{\textbf{Qualitative Results of Different 3D Control Strategies.} We use red circles to highlight regions where the 3D controls and our strategy significantly improve the interpolation quality.}
  \label{fig:vis_3dctrl}
  \vspace{-2.5pt}
\end{figure*}
\noindent\textbf{Setup.} We evaluate the effectiveness of the 3D-informed control design in PoseFuse3D via comparisons among 3D control strategies, including `VE', `VE+DN' and `VE+SE'. Specifically, VE refers to the visual encoding on the SMPL-X colored surface rendering, removing the SMPL-X encoder from the full design of our control model. VE+DN extends VE by incorporating depth and normal renderings, using two additional encoders identical to the one used in VE. VE+SE represents our proposed strategy in PoseFuse3D that directly encodes 3D information via the SMPL-X encoder and integrates it with VE. Experiments for these strategies use Wan2.1-I2V~\cite{wang2025wan} as the interpolation backbone. For efficiency, they are trained for 40K iterations.

To assess the necessity of 3D information, we compare against FCVG~\cite{zhu2024generative}, which is conditioned solely on 2D signals. For fair comparison, we create a variant of VE by replacing the backbone with SVD~\cite{blattmann2023svd}. We adapt SVD for interpolation with the temporal forward and reverse denoising path fusion strategy used in FCVG. This variant is referred to as VE-SVD for ease of analysis.

\noindent\textbf{Results.} We provide quantitative comparisons of 3D control strategies in Table~\ref{tab:main_table}. In comparisons between FCVG and VE-SVD, we find that adding 3D control improves the interpolation performance. VE-SVD outperforms FCVG across metrics with a more than \textbf{0.50 dB} increase on all the PSNRs, indicating the improvements on both whole-image and human-centric levels.  Moreover, our study highlights the importance of explicit 3D information. VE+DN and VE+SE, which incorporate depth and normal maps or direct SMPL-X information, outperform the simpler strategy VE. VE+DN and VE+SE show clear improvements in perceptual quality as reflected by the LPIPS metrics. Notably, our VE+SE strategy, which directly encodes information in 3D space, delivers the best performance, achieving the lowest LPIPS\textsubscript{bbox} of \textbf{0.0653} and the highest PSNR of \textbf{22.14 dB}. 

\noindent\textbf{Visualizations.} 
Besides our quantitative analysis, we perform a qualitative comparison of the 3D control strategies, as shown in Figure~\ref{fig:vis_3dctrl}.  We find that incorporating 3D controls better preserves human shape during interpolation.  Take the Tennis case in Figure~\ref{fig:vis_3dctrl} for example, methods with 3D control strategies interpolate the player's body close to the ground truth, whereas FCVG exhibits noticeable distortion. Moreover, our VE+SE, which directly encodes 3D information from SMPL-X, proves effective in handling occluded human motion. In both the Skateboarding ($1^{st}$ row) and Jumping ($2^{nd}$ row) cases in Figure~\ref{fig:vis_3dctrl}, we can observe that our VE+SE strategy produces plausible results for the occluded arms, demonstrating its advantage in complex scenarios.

\subsection{Benchmark Results}
\label{sec:bm}
\begin{table}[t]
\caption{\textbf{Comparisons with State-of-the-art Interpolation Methods.} 
}
\small
\label{tab:benchmark}
\centering
\setlength\tabcolsep{3pt}
\resizebox{0.8\textwidth}{!}{
\begin{tabular}{lccccccc}
% \toprule
\hline
\multirow{2}{*}{Methods}  & \multicolumn{7}{c}{Metrics} \\ \cmidrule(l){2-8}
& PSNR$\uparrow$  & PSNR\textsubscript{bbox}$\uparrow$ & PSNR\textsubscript{mask}$\uparrow$  & LPIPS$\downarrow$  & LPIPS\textsubscript{bbox}$\downarrow$  & LPIPS\textsubscript{mask}$\downarrow$ & HA$\uparrow$ \\ \midrule
GIMM-VFI~\cite{guo2025generalizable} & 20.29 & 16.36 & 14.93 & 0.1954  & 0.1187 & 0.0860 & 0.9146 \\ \midrule
GI~\cite{wang2024generative} & 15.81 & 13.04 & 12.03 & 0.3364  & 0.1672 & 0.1146 & 0.8954 \\
Wan2.1-KI (Ours) & 19.53 & 15.96 & 14.62 & 0.2081  & 0.1208 & 0.0868 & 0.9180\\ \midrule
FCVG~\cite{zhu2024generative} & 20.42 & 18.05 & 16.91 & 0.2100  & 0.0899 & 0.0606 & 0.9187 \\
PoseFuse3D-KI (Ours) & \textbf{22.27} & \textbf{19.49} & \textbf{18.24} & \textbf{0.1304}  & \textbf{0.0636} & \textbf{0.0450} & \textbf{0.9189}\\
\hline
\end{tabular}
}
\end{table}
\begin{figure*}[t]
  \vspace{-5pt}
  \centering
  \includegraphics[width=1.0\linewidth]{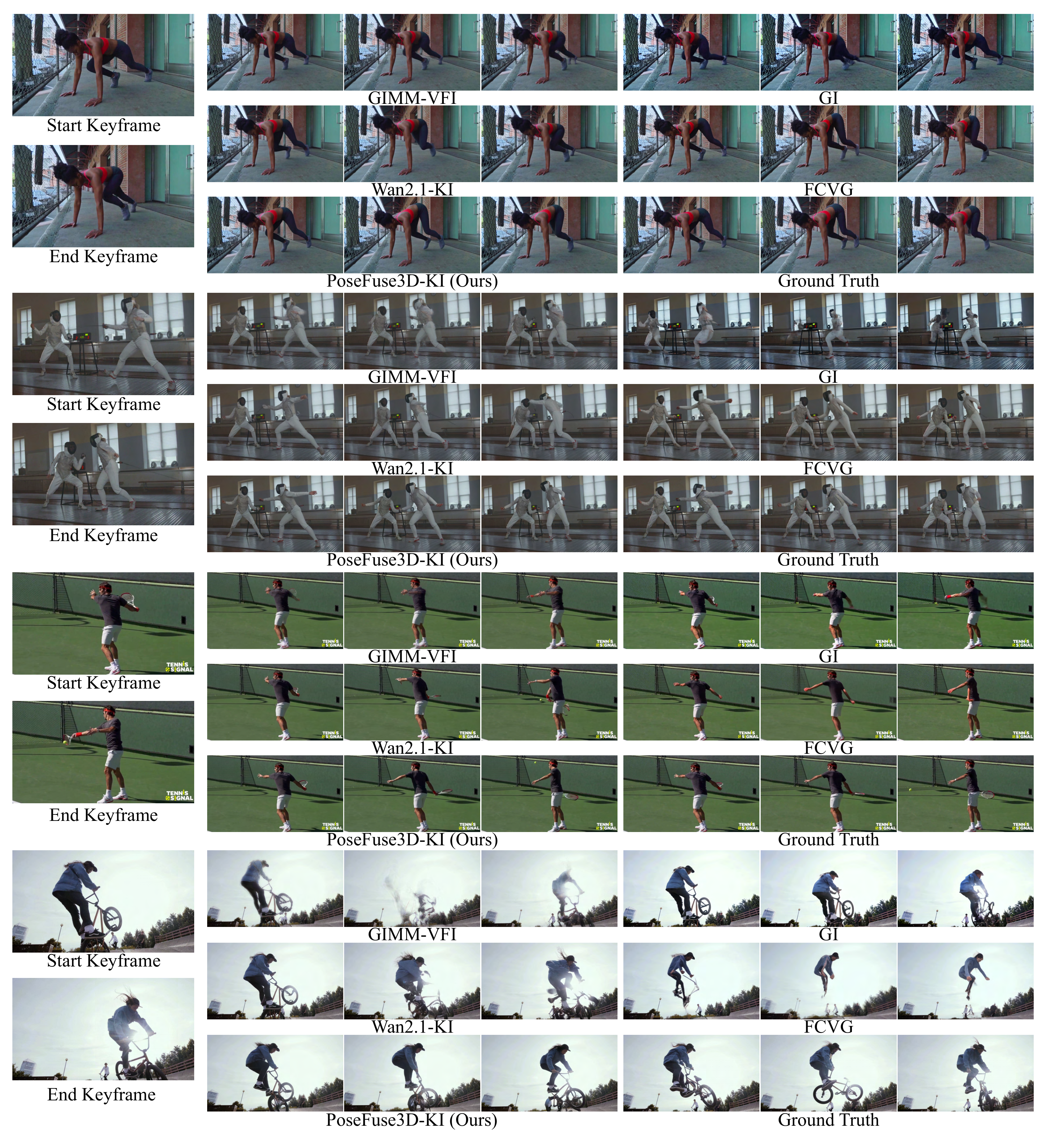}
  \caption{\textbf{Qualitative Comparisons with State-of-The-Art Methods.} }
  \label{fig:bm_res}
  \vspace{-8pt}
\end{figure*}
\noindent\textbf{Setup.} 
We compare PoseFuse3D-KI against several advanced interpolation methods on our CHKI-Video dataset. The main comparison is with FCVG~\cite{zhu2024generative}, which also enables intermediate control during interpolation. For broader coverage, we also include the keyframe interpolation method GI~\cite{wang2024generative} and the traditional video frame interpolation method GIMM-VFI~\cite{guo2025generalizable}. We also include Wan2.1-KI, an adaptation of Wan2.1~\cite{wang2025wan} for keyframe interpolation, following the strategy in Sec.~\ref{sec:adaptevdm}. Since not all baselines support ground truth controls, we additionally compute the Human Anatomy (HA)~\cite{zheng2025vbench} score to assess the quality of human synthesis and support fair comparison.

\noindent\textbf{Results.} 
We report quantitative results of PoseFuse3D‑KI on the CHKI‑Video benchmark in Table~\ref{tab:benchmark}. Our method delivers state‑of‑the‑art performance on human‑centric keyframe interpolation. On whole‑image metrics, it boosts PSNR by \textbf{1.85 dB} and lowers LPIPS by \textbf{0.0796} in comparison with the state-of-the-art method FCVG~\cite{zhu2024generative}.  Crucially, it also outperforms on human-centric metrics compared with other methods, achieving a PSNR\textsubscript{bbox} of \textbf{19.49 dB}, an LPIPS\textsubscript{mask} of \textbf{0.045}, and an HA score of \textbf{0.9189}. This indicates that our method produces plausible, high‑fidelity human interpolations that closely follow the ground‑truth dynamics, demonstrating the effectiveness of our method.

\noindent\textbf{Visualizations.}
For qualitative evaluation, we qualitatively compare PoseFuse3D-KI with other advanced methods, as illustrated in Figure~\ref{fig:bm_res}. Our approach delivers more accurate human interpolations, faithfully following real-world motions and preserving body shape. For example, in the $2^{nd}$ `Fencing' case and $4^{th}$ `Stunt Bike' case, only PoseFuse3D‑KI correctly interpolates leg and arm movements while maintaining consistent shapes. Moreover, our method naturally handles the occluded human motion well with 3D-informed control. We observe that our method interpolates correct spatial positions for the occluded legs ($1^{st}$ case) and arms ($3^{nd}$ case), achieving significant improvements over FCVG. Furthermore, although the control-free keyframe interpolation methods GI and Wan2.1-KI occasionally generate undistorted humans, they often generate implausible motion that violates real‑world dynamics, as observed in $1^{st}$, $2^{nd}$, and $4^{th}$ cases.

\subsection{Scalability Across Temporal Gaps}
\label{sec:temporal_gaps}
\begin{table}[t]
\caption{\textbf{Results across Temporal Gaps.} 
}
\small
\label{tab:temporal_gap}
\centering
\setlength\tabcolsep{5pt}
\resizebox{0.8\textwidth}{!}{
\begin{tabular}{lcccccc}
% \toprule
\hline
\multirow{2}{*}{Temporal Gap}  & \multicolumn{6}{c}{Metrics} \\ \cmidrule(l){2-7}
& PSNR$\uparrow$  & PSNR\textsubscript{bbox}$\uparrow$ & PSNR\textsubscript{mask}$\uparrow$  & LPIPS$\downarrow$  & LPIPS\textsubscript{bbox}$\downarrow$  & LPIPS\textsubscript{mask}$\downarrow$  \\ \midrule
24-frame & 23.86 & 21.64 & 20.70  & 0.1144  & 0.0508 & 0.0332 \\
48-frame & 21.46 & 19.48 & 18.57  & 0.1566  & 0.0644 & 0.0432 \\
64-frame & 20.37 & 18.34 & 17.45  & 0.1882  & 0.0755 & 0.0514\\
96-frame & 19.44 & 17.50 & 16.66  & 0.2208  & 0.0836 & 0.0577\\
% \bottomrule
\hline 			
\end{tabular}
}
\end{table}
\noindent\textbf{Setup.} 
We evaluate the scalability of PoseFuse3D-KI in interpolating frames across different temporal gaps. The temporal gap is defined as the number of frames between the input keyframes, ranging from 24 to 96 frames. Evaluations are conducted on the CHKI-Video test set using PSNR and LPIPS as quantitative metrics.

\noindent\textbf{Results.} 
We provide quantitative results of PoseFuse3D-KI across different temporal gaps in Table~\ref{tab:temporal_gap}. As a common trend observed in most interpolation methods, the interpolation performance gradually declines as the interval between keyframes increases. Nevertheless, PoseFuse3D-KI maintains strong performance over a wide range of temporal settings. Notably, it achieves a PSNR of \textbf{23.86 dB} and an LPIPS of \textbf{0.1144} at 24 frames, and a PSNR of \textbf{21.46 dB} and an LPIPS of \textbf{0.1566} at 48 frames. Even at a large gap of 96 frames, where the task becomes substantially more challenging, PoseFuse3D-KI remains competitive, achieving a PSNR of \textbf{19.44 dB} and an LPIPS of \textbf{0.2208}. These results demonstrate the robustness of PoseFuse3D-KI under long-range controllable interpolation.

\subsection{In-the-wild Interpolation Results}
\label{sec:itw_interp_results}
Our PoseFuse3D-KI framework can be readily applied to interpolate in-the-wild human-centric keyframes \textbf{without ground-truth control signals}.

\noindent\textbf{Setup.} 
We assess the in-the-wild interpolation capability of PoseFuse3D-KI on the CHKI-Video dataset by discarding the ground-truth annotations. The comparison is conducted against the state-of-the-art method FCVG~\cite{zhu2024generative}. Since ground-truth controls are unavailable in this setting, we employ a simple strategy that linearly interpolates human body joints to generate intermediate control signals. Details of this pipeline are provided in the supplementary material. For performance measurement, we compute the same metrics as described in Sec.~\ref{sec:bm}.

\noindent\textbf{Results.} 
We report quantitative results of PoseFuse3D-KI under the in-the-wild interpolation setting in Table~\ref{tab:linear_benchmark}. Our method achieves state-of-the-art performance and delivers significant improvements over FCVG~\cite{zhu2024generative}. On whole-image metrics, it boosts PSNR by \textbf{4.5\%} and reduces LPIPS by \textbf{22.1\%} compared with FCVG. More importantly, PoseFuse3D-KI attains the best results on perceptual human-centric metrics, with an LPIPS\textsubscript{mask} of \textbf{0.0859} and an HA score of \textbf{0.9289}. These results demonstrate that PoseFuse3D-KI effectively handles in-the-wild interpolation.

\subsection{Ablation Study}
\label{sec:ablation}
In this section, we ablate the visual encoding and fusion modules of PoseFuse3D in Table~\ref{tab:ablation}. For efficiency, we use SVD as the backbone and process video clips into 9 consecutive frames of $256\times256$. We use 3× temporally downsampled videos from the CHKI-Video test set for evaluation. 

\noindent\textbf{3D Visual Encoding.}
Our visual encoding module includes two convolutional encoders for 2D and 3D control maps, respectively. We denote the variant including this entire module as `Dual Conv-Enc', and the one using only the 2D encoder as `Conv-Enc (2D)'. Removing the 3D visual encoding leads to a performance drop of 0.26 dB on both PSNR\textsubscript{bbox} and PSNR\textsubscript{mask}, highlighting the importance of 3D visual encoding.

\noindent\textbf{Fusion Module.} 
In PoseFuse3D, condition features are fused through a carefully designed fusion module. To validate its effectiveness, we replace it with a simple summation operation~\cite{zhu2024champ}, denoted as `Sum' in Table~\ref{tab:ablation}. This change leads to a significant performance drop, particularly in perceptual quality, with an increase of 0.0033 in LPIPS\textsubscript{mask}. These results demonstrate the fusion module's contribution to providing informative control for high-quality interpolation. 

\noindent\textbf{Window-Partition Strategy.}
PoseFuse3D employs a cross-attention layer with a shifted window-partition strategy to fuse features across neighboring windows. To validate this design, we remove the window partitioning, denoted as Non-WP. This results in notable drops of 0.11 dB in both PSNR\textsubscript{bbox} and PSNR\textsubscript{mask}, indicating that the window-partition strategy enhances controlled interpolation.

\noindent\textbf{Temporal Attention in Fusion Module.} 
To justify the efficacy of the temporal self-attention (TSA) layer in the fusion module, we conduct experiments excluding the TSA layer (\textit{Non-TSA}). This removal causes increases of 0.0022 and 0.0017 in LPIPS\textsubscript{bbox} and LPIPS\textsubscript{mask}, demonstrating the crucial role of the temporal self-attention layers in the fusion module.

\begin{table}[t]
\caption{\textbf{Comparisons with FCVG on In-the-wild Interpolation.} 
}
\small
\label{tab:linear_benchmark}
\centering
\setlength\tabcolsep{3pt}
\resizebox{\textwidth}{!}{
\begin{tabular}{lccccccc}
% \toprule
\hline
\multirow{2}{*}{Methods}  & \multicolumn{7}{c}{Metrics} \\ \cmidrule(l){2-8}
& PSNR$\uparrow$  & PSNR\textsubscript{bbox}$\uparrow$ & PSNR\textsubscript{mask}$\uparrow$  & LPIPS$\downarrow$  & LPIPS\textsubscript{bbox}$\downarrow$  & LPIPS\textsubscript{mask}$\downarrow$ & HA$\uparrow$ \\ \midrule
FCVG~\cite{zhu2024generative} & 18.47 & 15.31 & 14.08  & 0.2607  & 0.1321 & 0.0921 & 0.9284 \\
PoseFuse3D-KI (Ours) & \textbf{19.30} & \textbf{15.75} & \textbf{14.46}  & \textbf{0.2031}  & \textbf{0.1194} & \textbf{0.0859} & \textbf{0.9289}\\
% \bottomrule
\hline
\end{tabular}
}
\end{table}

\begin{table}[t]
\caption{\textbf{Ablation Study on Visual Encoding and Fusion Module.} 
}
\label{tab:ablation}
\centering
\setlength\tabcolsep{3pt}
\resizebox{\textwidth}{!}{
\begin{tabular}{cccccccc}
\toprule
\multicolumn{2}{c}{Model Variants} & \multicolumn{6}{c}{Evaluation Metrics} \\ \cmidrule(l){1-2} \cmidrule(l){3-8}
Visual Encoding & Fusion Module & PSNR & PSNR\textsubscript{bbox}$\uparrow$ & PSNR\textsubscript{mask}$\uparrow$  & LPIPS$\downarrow$  & LPIPS\textsubscript{bbox}$\downarrow$  & LPIPS\textsubscript{mask}$\downarrow$ \\ \midrule
Conv-Enc (2D) &  N.A. &  20.29 & 16.94 & 16.10  & 0.1990  & 0.1096 & 0.0836 \\
Dual Conv-Enc &  Sum & 20.50 & 17.20 & 16.36  & 0.1953  & 0.1059 & 0.0806 \\
Dual Conv-Enc & Non-TSA &  20.30 & 16.96 & 16.12  & 0.1956  & 0.1053 & 0.0790 \\
Dual Conv-Enc & Non-WP &  20.46 & 17.19 & 16.37  & 0.1938  & 0.1038 & 0.0775 \\
Dual Conv-Enc & Full &  \textbf{20.55} & \textbf{17.30} & \textbf{16.48}  & \textbf{0.1927}  & \textbf{0.1031} & \textbf{0.0773} \\

\bottomrule
\end{tabular}
}
\vspace{-10pt}
\end{table}

\section{Conclusion}
We propose PoseFuse3D‑KI, a controllable human‑centric keyframe interpolation framework powered by our novel 3D‑informed control model, PoseFuse3D. PoseFuse3D embeds rich spatial geometry from 3D human signals together with 2D poses into a unified control feature, enabling the generation of more plausible and realistic intermediate frames. For evaluation, we construct a CHKI-Video dataset with comprehensive human-centric annotations. Extensive experiments on the benchmark demonstrate that PoseFuse3D-KI outperforms previous interpolation methods with a 9\% improvement in PSNR and a 38\% reduction in LPIPS.

% ---- Bibliography ----

\clearpage
\noindent\textbf{Acknowledgement.} 
This study is supported under the RIE2020 Industry Alignment Fund Industry Collaboration Projects (IAF-ICP) Funding Initiative, as well as cash and in-kind contribution from the industry partner(s).
\bibliography{main}

% ---- Checklist ----
%%%%%%%%%%%%%%%%%%%%%%%%%%%%%%%%%%%%%%%%%%%%%%%%%%%%%%%%%%%%

\newpage
\section*{NeurIPS Paper Checklist}

%%% BEGIN INSTRUCTIONS %%%
The checklist is designed to encourage best practices for responsible machine learning research, addressing issues of reproducibility, transparency, research ethics, and societal impact. Do not remove the checklist: {\bf The papers not including the checklist will be desk rejected.} The checklist should follow the references and follow the (optional) supplemental material.  The checklist does NOT count towards the page
limit. 

Please read the checklist guidelines carefully for information on how to answer these questions. For each question in the checklist:
\begin{itemize}
    \item You should answer \answerYes{}, \answerNo{}, or \answerNA{}.
    \item \answerNA{} means either that the question is Not Applicable for that particular paper or the relevant information is Not Available.
    \item Please provide a short (1–2 sentence) justification right after your answer (even for NA). 
   % \item {\bf The papers not including the checklist will be desk rejected.}
\end{itemize}

{\bf The checklist answers are an integral part of your paper submission.} They are visible to the reviewers, area chairs, senior area chairs, and ethics reviewers. You will be asked to also include it (after eventual revisions) with the final version of your paper, and its final version will be published with the paper.

The reviewers of your paper will be asked to use the checklist as one of the factors in their evaluation. While "\answerYes{}" is generally preferable to "\answerNo{}", it is perfectly acceptable to answer "\answerNo{}" provided a proper justification is given (e.g., "error bars are not reported because it would be too computationally expensive" or "we were unable to find the license for the dataset we used"). In general, answering "\answerNo{}" or "\answerNA{}" is not grounds for rejection. While the questions are phrased in a binary way, we acknowledge that the true answer is often more nuanced, so please just use your best judgment and write a justification to elaborate. All supporting evidence can appear either in the main paper or the supplemental material, provided in appendix. If you answer \answerYes{} to a question, in the justification please point to the section(s) where related material for the question can be found.

IMPORTANT, please:
\begin{itemize}
    \item {\bf Delete this instruction block, but keep the section heading ``NeurIPS Paper Checklist"},
    \item  {\bf Keep the checklist subsection headings, questions/answers and guidelines below.}
    \item {\bf Do not modify the questions and only use the provided macros for your answers}.
\end{itemize}

%%% END INSTRUCTIONS %%%

\begin{enumerate}

\item {\bf Claims}
    \item[] Question: Do the main claims made in the abstract and introduction accurately reflect the paper's contributions and scope?
    \item[] Answer: \answerYes{} % Replace by \answerYes{}, \answerNo{}, or \answerNA{}.
    \item[] Justification: We have made clear claims in the abstract and introduction that accurately reflect our paper’s contributions to the studied problem. Please refer to the abstract and introduction for more detailed information.

\item {\bf Limitations}
    \item[] Question: Does the paper discuss the limitations of the work performed by the authors?
    \item[] Answer: \answerYes{} % Replace by \answerYes{}, \answerNo{}, or \answerNA{}.
    \item[] Justification: We have discussions on the limitations in the supplementary materials.

\item {\bf Theory assumptions and proofs}
    \item[] Question: For each theoretical result, does the paper provide the full set of assumptions and a complete (and correct) proof?
    \item[] Answer: \answerNA{} % Replace by \answerYes{}, \answerNo{}, or \answerNA{}.
    \item[] Justification: Our work does not include theoretical results.

    \item {\bf Experimental result reproducibility}
    \item[] Question: Does the paper fully disclose all the information needed to reproduce the main experimental results of the paper to the extent that it affects the main claims and/or conclusions of the paper (regardless of whether the code and data are provided or not)?
    \item[] Answer: \answerYes{} % Replace by \answerYes{}, \answerNo{}, or \answerNA{}.
    \item[] Justification: We provide the implementation details in the main text and training hyperparameters in the supplementary materials.

\item {\bf Open access to data and code}
    \item[] Question: Does the paper provide open access to the data and code, with sufficient instructions to faithfully reproduce the main experimental results, as described in supplemental material?
    \item[] Answer: \answerNo{} % Replace by \answerYes{}, \answerNo{}, or \answerNA{}.
    \item[] Justification: We are unable to provide our code upon submission, but releasing the code to the public in the future is our plan.

\item {\bf Experimental setting/details}
    \item[] Question: Does the paper specify all the training and test details (e.g., data splits, hyperparameters, how they were chosen, type of optimizer, etc.) necessary to understand the results?
    \item[] Answer: \answerYes{} % Replace by \answerYes{}, \answerNo{}, or \answerNA{}.
    \item[] Justification: We provide the general training information and evaluation settings in the main text. The full implementation details and training hyperparameters are presented in the supplementary materials.

\item {\bf Experiment statistical significance}
    \item[] Question: Does the paper report error bars suitably and correctly defined or other appropriate information about the statistical significance of the experiments?
    \item[] Answer: \answerNo{} % Replace by \answerYes{}, \answerNo{}, or \answerNA{}.
    \item[] Justification: This paper does not report error bars.

\item {\bf Experiments compute resources}
    \item[] Question: For each experiment, does the paper provide sufficient information on the computer resources (type of compute workers, memory, time of execution) needed to reproduce the experiments?
    \item[] Answer: \answerYes{} % Replace by \answerYes{}, \answerNo{}, or \answerNA{}.
    \item[] Justification: We provide the type and the number of GPUs used in our experiments.
    
\item {\bf Code of ethics}
    \item[] Question: Does the research conducted in the paper conform, in every respect, with the NeurIPS Code of Ethics \url{https://neurips.cc/public/EthicsGuidelines}?
    \item[] Answer: \answerYes{} % Replace by \answerYes{}, \answerNo{}, or \answerNA{}.
    \item[] Justification: We carefully follow the NeurIPS Code of Ethics.

\item {\bf Broader impacts}
    \item[] Question: Does the paper discuss both potential positive societal impacts and negative societal impacts of the work performed?
    \item[] Answer: \answerYes{} % Replace by \answerYes{}, \answerNo{}, or \answerNA{}.
    \item[] Justification: We discuss the potential impacts in the supplementary materials.
    
\item {\bf Safeguards}
    \item[] Question: Does the paper describe safeguards that have been put in place for responsible release of data or models that have a high risk for misuse (e.g., pretrained language models, image generators, or scraped datasets)?
    \item[] Answer: \answerNA{} % Replace by \answerYes{}, \answerNo{}, or \answerNA{}.
    \item[] Justification: The paper poses no such risks.

\item {\bf Licenses for existing assets}
    \item[] Question: Are the creators or original owners of assets (e.g., code, data, models), used in the paper, properly credited and are the license and terms of use explicitly mentioned and properly respected?
    \item[] Answer: \answerYes{} % Replace by \answerYes{}, \answerNo{}, or \answerNA{}.
    \item[] Justification:  In our paper, we have cited the related papers and dataset sources.

\item {\bf New assets}
    \item[] Question: Are new assets introduced in the paper well documented and is the documentation provided alongside the assets?
    \item[] Answer: \answerNo{} % Replace by \answerYes{}, \answerNo{}, or \answerNA{}.
    \item[] Justification: Upon submission, we do not provide the source code and the dataset. But we will make them public in the future after acceptance.

\item {\bf Crowdsourcing and research with human subjects}
    \item[] Question: For crowdsourcing experiments and research with human subjects, does the paper include the full text of instructions given to participants and screenshots, if applicable, as well as details about compensation (if any)? 
    \item[] Answer: \answerNA{} % Replace by \answerYes{}, \answerNo{}, or \answerNA{}.
    \item[] Justification: Our paper does not involve crowdsourcing nor research with human subjects.

\item {\bf Institutional review board (IRB) approvals or equivalent for research with human subjects}
    \item[] Question: Does the paper describe potential risks incurred by study participants, whether such risks were disclosed to the subjects, and whether Institutional Review Board (IRB) approvals (or an equivalent approval/review based on the requirements of your country or institution) were obtained?
    \item[] Answer: \answerNA{} % Replace by \answerYes{}, \answerNo{}, or \answerNA{}.
    \item[] Justification: Our paper does not involve crowdsourcing nor research with human subjects.

\item {\bf Declaration of LLM usage}
    \item[] Question: Does the paper describe the usage of LLMs if it is an important, original, or non-standard component of the core methods in this research? Note that if the LLM is used only for writing, editing, or formatting purposes and does not impact the core methodology, scientific rigorousness, or originality of the research, declaration is not required.
    %this research? 
    \item[] Answer: \answerNA{} % Replace by \answerYes{}, \answerNo{}, or \answerNA{}.
    \item[] Justification: The core method development in this research does not involve LLMs as any important, original, or non-standard components.

\end{enumerate}

% ---- Supplementary ----
\clearpage
\section{Appendix}
In this section, as referenced in the main text, we first provide detailed descriptions of our method and dataset in Sec.~\ref{sec:wanki_sp}, Sec.~\ref{sec:chkiv_d}, and Sec.~\ref{sec:impld}. We then present additional experiments on our approach, including in-the-wild interpolation results(Sec.~\ref{sec:itw_interp}), extended benchmarking on the FCVG~\cite{zhu2024generative} test set (Sec.~\ref{sec:add_bm}), user study~\ref{sec:user_study}, and ablation studies on the control signals (Sec.~\ref{sec:asve}).
as well as the SMPL encoder (Sec.~\ref{sec:ase}). 
We also provide supplementary visualizations in Sec.~\ref{sec:av} and analyze model runtime in Sec.~\ref{sec:runtime}. Finally, we discuss the limitations and broader impact of our method in Sec.~\ref{sec:lim} and Sec.~\ref{sec:bi}.

\subsection{Wan2.1 for Keyframe Interpolation}
\label{sec:wanki_sp}
We adapt the Wan2.1~\cite{wang2025wan} Image-to-Video model (14B, 480P) as the keyframe interpolator Wan2.1-KI along a single temporal forward diffusion path. Specifically, we insert zero-padding between the input keyframes to construct a full-length video sequence. This sequence is then encoded into latents using the VAE encoder of Wan2.1. The resulting latent representation is concatenated with a noisy latent and a latent mask, and passed to the denoising network for prediction. In parallel, the input keyframes are encoded using the image CLIP encoder to produce condition tokens, which guide the denoising process through cross-attention mechanisms. To accommodate changes in both the latent inputs and attention layer inputs, we perform parameter-efficient LoRA fine-tuning on the input embedding layer and on the value and output projection matrices of the attention layers.

\subsection{CHKI-Video: Detailed Construction Stages}
Our dataset is specifically designed for the Controllable Human-centric Keyframe Interpolation (CHKI) task, emphasizing complex human motions, human-centric annotations, and distant keyframe inputs. To this end, we carefully control the data collection and annotation process, sourcing videos from task-relevant datasets~\cite{chen2024sportsslomo} or the internet using specifically curated keywords, rather than relying on large-scale collections~\cite{ju2024miradata, wang2024humanvid} that may include irrelevant content.

\label{sec:chkiv_d}
\noindent\textbf{Stage 1: Dataset Collection.} 
We begin by collecting video clips from SportSlomo~\cite{chen2024sportsslomo}, which are temporally downsampled to 60 fps due to the large motions typically present in sports scenarios, making them more challenging for keyframe interpolation.
To enhance dataset diversity versus reality~\cite{yu2023dataset}, we additionally crawl high-quality stock videos from the Pexels website. We compile a list of keywords representing fundamental human movements such as `Walking’, `Kicking’, `Throwing’, `Catching’, and `Climbing’, to cover a broad range of human activities. For each keyword, we collect 100 unique videos with resolutions above 720p and durations under 30 seconds. These keywords are grouped into three motion categories: arm motion, leg motion, and general motion, ensuring balanced labels for subsequent train-test splitting. To match the motion characteristics of the SportSlomo videos, we downsample the collected stock videos based on their optical flow scores, ensuring the flow score distributions are aligned.

\noindent\textbf{Stage 2: Pre-annotation Processing.}
To ensure the quality of the collected videos, we use DOVER~\cite{wu2023dover} to obtain both technical and overall quality scores, and compute brightness change scores between adjacent frames. Videos falling below the bottom 5th percentile in any of these metrics are filtered out. Given the importance of accurate human detection for downstream keypoint and SMPL-X annotation, we design a robust detection pipeline. We combine Grounding-DINO~\cite{liu2024groundingdino} with SAM2~\cite{ravi2024sam2} to achieve reliable human detection. For challenging sports scenes, we prioritize videos with prominent foreground humans and relatively static or blurred backgrounds, striking a balance between annotation complexity and scenario diversity. Additionally, we exclude videos containing more than three people or fewer than 20 frames to maintain clean motion patterns and ensure sufficient temporal coverage. All sports videos are manually reviewed to verify compliance with these criteria and to confirm the accuracy of the human detections.

\noindent\textbf{Stage 3: Human-centric Annotation.}
We perform frame-wise human-centric annotations for all video clips based on the detections in the previous stage.
First, we use Sapiens~\cite{khirodkar2024sapiens} to estimate whole-body keypoints. To ensure the dataset remains strictly human-centric, we perform whole-body detection based on these keypoints. Specifically, we extract the keypoints into DWPose to better define the human figure.  We merge all head keypoints into a single point, as significant motion rarely occurs in that region. A whole-body detection is considered valid if it contains fewer than three invalid keypoints, using a keypoint score threshold of 0.3. We further filter video clips to retain only those with more than 20 consecutive valid frames. Finally, we apply SMPLer-X~\cite{cai2023smpler}, which provides high re-projection accuracy, to fit detailed SMPL-X models to each frame and generate reliable 3D body parameters. 

\subsection{Implementation Details}
\label{sec:impld}
We fine-tune the entire PoseFuse3D-KI framework in an end-to-end manner using the AdamW optimizer with a learning rate of $8\times10^{-5}$. The fine-tuning is applied to our 3D-informed control model, PoseFuse3D, with additional LoRA adaptation on the input patch embeddings, as well as the value and output projections of the VDM’s attention modules. Both the LoRA rank and LoRA alpha are set to 32. For implementation, we leverage Fully Sharded Data Parallel (FSDP) across 4 GPUs.

\begin{figure*}[t]
  \vspace{-10pt}
  \centering
  \includegraphics[width=1.0\linewidth]{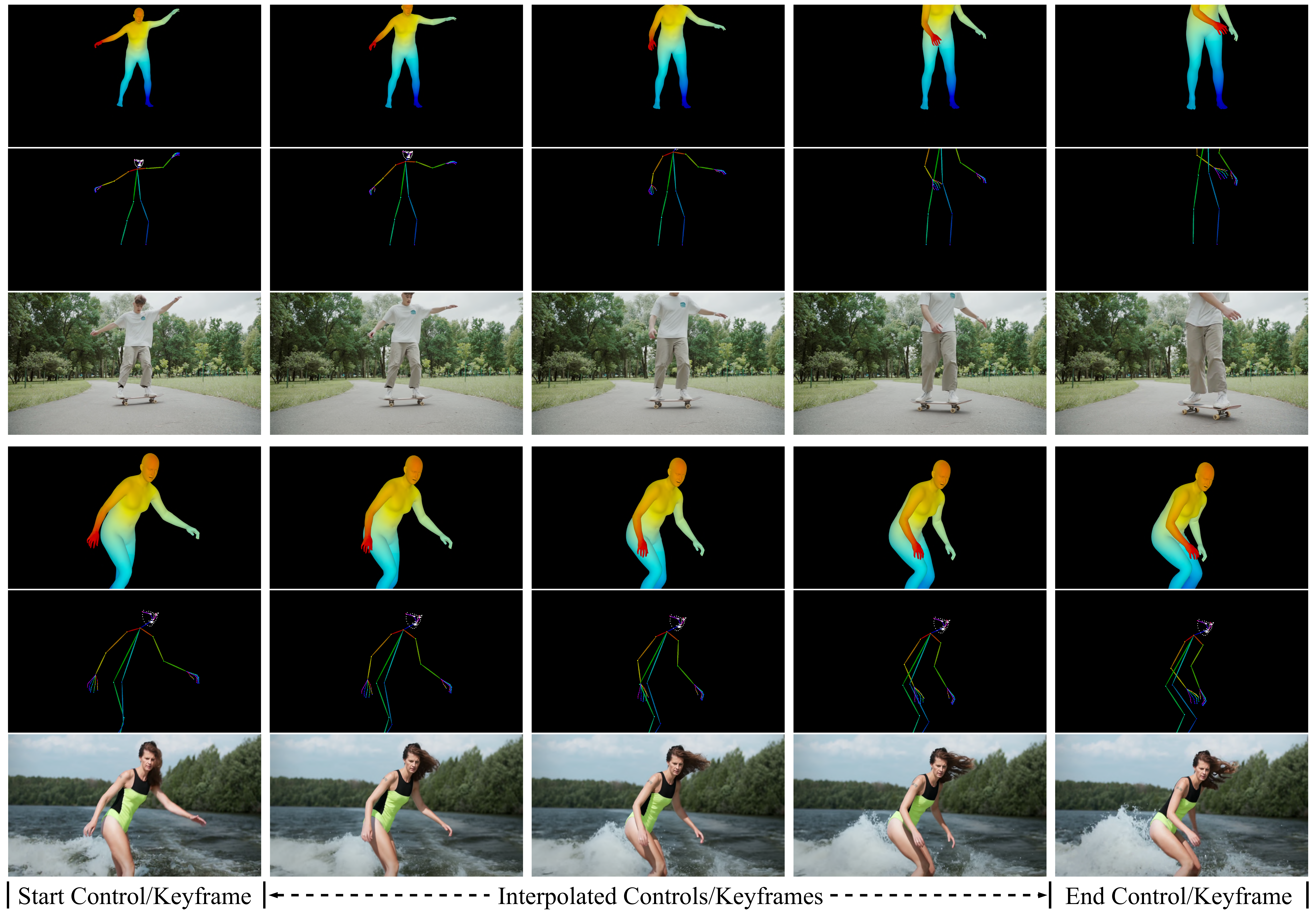}
  \caption{\textbf{Qualitative Results of In-the-wild Control and Keyframe Interpolation.} }
  \label{fig:supp_itw}
  \vspace{-8pt}
\end{figure*}
\subsection{In-the-wild Interpolation}
\label{sec:itw_interp}
Our PoseFuse3D-KI framework can be readily applied to interpolate in-the-wild human-centric keyframes. In this subsection, we present a simple pipeline that uses linear interpolation for human body joints. 
Given a human-centric keyframe pair $I_0, I_N$, we first employ a 3D human model estimator, such as SMPLer-X~\cite{cai2023smpler}, to fit SMPL-X models~\cite{pavlakos2019smplx} for each keyframe input. Leveraging the strong human body priors from SMPL-X, we linearly interpolate the SMPL-X parameters to generate intermediate 3D human models, which serve as control signals for interpolation. We then extract 2D DWPose keypoints from the 2D projections of the interpolated SMPL-X models. With these steps, all necessary guidance inputs for PoseFuse3D-KI are prepared and can be directly used for keyframe interpolation. Figure~\ref{fig:supp_itw} shows the results of the interpolated SMPL-X models and the corresponding video frames. 
While this pipeline offers a straightforward and efficient approach for generating intermediate poses, it is limited in its ability to handle complex human motion. In particular, it struggles to capture realistic dynamics such as acceleration or multi-step actions. To address these limitations, we could explore text-to-motion models~\cite{zhao2024dartcontrol} as part of future work, which generate intermediate SMPL-X models from textual descriptions, offering a more flexible and semantically rich alternative to linear interpolation.

\subsection{Evaluation on Additional Benchmark}
\label{sec:add_bm}
\begin{table}[t]
\caption{\textbf{Benchmark Results on FCVG-Test-HC.} 
}
\small
\label{tab:benchmark_fcvgt}
\centering
\setlength\tabcolsep{3pt}
\resizebox{0.8\textwidth}{!}{
\begin{tabular}{lccccccc}
% \toprule
\hline
\multirow{2}{*}{Methods}  & \multicolumn{7}{c}{Metrics} \\ \cmidrule(l){2-8}
& PSNR$\uparrow$  & PSNR\textsubscript{bbox}$\uparrow$ & PSNR\textsubscript{mask}$\uparrow$  & LPIPS$\downarrow$  & LPIPS\textsubscript{bbox}$\downarrow$  & LPIPS\textsubscript{mask}$\downarrow$ & HA$\uparrow$ \\ \midrule
GIMM-VFI~\cite{guo2025generalizable} & 23.61 & 21.25 & 20.28  & 0.1324 & 0.0759 & 0.0587 & \textbf{0.9459} \\ \midrule
GI~\cite{wang2024generative} & 17.21 & 15.20 & 14.36  & 0.2701  & 0.1422 & 0.1045 & 0.9438 \\
Wan2.1-KI (Ours) & 21.50 & 19.03 & 18.17  & 0.1553  & 0.0915 & 0.0704 & 0.9312 \\ \midrule
FCVG~\cite{zhu2024generative} & 22.49 & 21.03 & 20.69  & 0.1738  & 0.0734 & 0.0493 & 0.9241 \\
PoseFuse3D-KI (Ours) & \textbf{24.84} & \textbf{22.97} & \textbf{22.26}  & \textbf{0.0915}  & \textbf{0.0460} & \textbf{0.0340} & 0.9245\\
% \bottomrule
\hline
\end{tabular}
}
\end{table}
\begin{figure*}[t]
  \vspace{-10pt}
  \centering
  \includegraphics[width=1.0\linewidth]{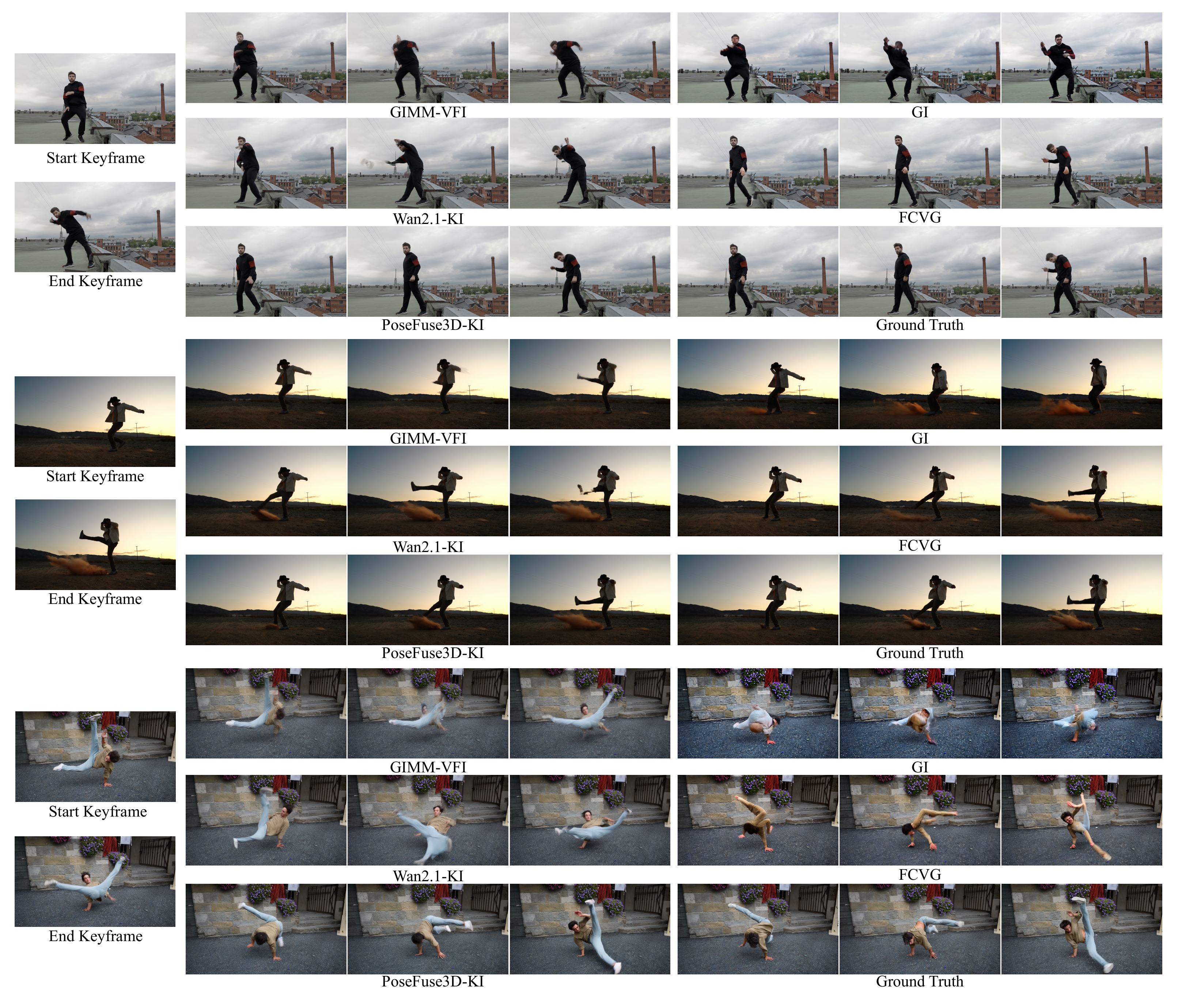}
  \caption{\textbf{Qualitative Comparisons on FCVG-Test-HC.} }
  \label{fig:supp_bm}
  \vspace{-8pt}
\end{figure*}

\noindent\textbf{Setup.} 
We conduct an additional evaluation on the test set from FCVG~\cite{zhu2024generative}. We first perform human detection and extract human-centric video clips from the test set. The extracted videos are then annotated using the same processing pipeline detailed in Sec.~\ref{sec:chkiv_d}. This results in FCVG-Test-HC, a curated human-centric subset of 54 clips suitable for CHKI benchmarking. The FCVG-Test-HC benchmark is relatively easier than CHKI-Video, primarily consisting of human-centric clips with limited motion rather than more challenging scenarios such as sports and dancing.
Other benchmarking settings follow those described in the main paper.

\noindent\textbf{Results.}
We present quantitative comparisons on the FCVG-Test-HC benchmark in Table~\ref{tab:benchmark_fcvgt}. Our PoseFuse3D-KI outperforms other methods for human-centric keyframe interpolation. Compared with the previous state-of-the-art method FCVG~\cite{zhu2024generative}, our method achieves a 7.6\% improvement in PSNR\textsubscript{mask} and a 31\% reduction in LPIPS\textsubscript{mask}. 
We observe that all methods achieve higher PSNR and lower LPIPS scores on the FCVG-Test-HC benchmark compared to the CHKI-Video dataset, indicating that FCVG-Test-HC is an easier benchmark for interpolation. This aligns with our earlier observation during dataset construction, where FCVG-Test-HC primarily consists of human-centric clips with limited motion. 
Interestingly, methods with fewer learned priors tend to achieve higher HA scores in this setting. For instance, the traditional interpolation method GIMM-VFI~\cite{guo2025generalizable} records the highest Human Anatomy (HA) score. This is likely because such methods rely more heavily on the input keyframes. While this reliance leads to motion artifacts under large movement, it better preserves human textures from inputs when the motion between keyframes is small.

\noindent\textbf{Visualizations.} We qualitatively compare PoseFuse3D-KI with other advanced methods on the FCVG-Test-HC benchmark, as shown in Figure~\ref{fig:supp_bm}. Consistent with our findings in the Benchmark Results section of the main paper, our method achieves robust human-centric interpolation, accurately follows real-world dynamics, and effectively preserves human body shape. For instance, our method generates plausible interpolations of complex body movements while maintaining the correct leg structure and posture in the last `Breaking Dance' case.

\subsection{User Study}
\label{sec:user_study}
\begin{table}[t]
\caption{\textbf{User Study.} 
}
\small
\label{tab:user_study}
\centering
\setlength\tabcolsep{3pt}
\resizebox{0.8\textwidth}{!}{
\begin{tabular}{lcccc}
\toprule
% \hline
Methods & GIMM-VFI~\cite{guo2025generalizable}  & GI~\cite{wang2024generative} & FCVG~\cite{zhu2024generative}  & PoseFuse3D-KI (Ours)  \\ \midrule
User Preference (\%) & 1.75 & 5.13 & 6.63 & \textbf{86.50} \\
\bottomrule
% \hline 			
\end{tabular}
}
\end{table}
\noindent\textbf{Setup.}
To further assess the perceptual quality of our controllable human-centric interpolation, we conducted a user study involving 20 interpolation scenarios sampled from both the CHKI-Video test set and FCVG-Test-HC. We compared PoseFuse-KI with three representative state-of-the-art methods: GIMM-VFI~\cite{guo2025generalizable}, GI~\cite{wang2024generative}, and FCVG~\cite{zhu2024generative}. A total of 40 participants were invited to take part in the study, where each participant was asked to select their preferred interpolation result. 

\noindent\textbf{Results.}
We report the user preference of the study in Table~\ref{tab:user_study}. Our method received a strong majority of user preferences (86.5\%), consistently outperforming all baselines. These results further demonstrate the perceptual effectiveness of PoseFuse3D-KI.

\subsection{Ablation Study on Visual Encoding}
\label{sec:asve}
The core of our framework is the control module, PoseFuse3D. As detailed in the Method section of the main paper, it includes encoding visualizations from both SMPL-X and DWPose~\cite{yang2023dwpose}. To evaluate the importance of encoding these visualizations and explore whether 2D visual cues can be omitted, we conduct an ablation study on the visual encoding component of PoseFuse3D.

\noindent\textbf{Necessity of Encoding Visualizations.} 
In PoseFuse3D, we encode visualizations of control signals. Since they preserve natural pixel-level alignment with the video latent, thereby providing direct control signals on the pixel plane. To assess its necessity, we ablate all visual encoding components in PoseFuse3D and rely solely on the SMPL-X encoded information as the control representation. We refer to this variant as `Non-Vis'. In Table~\ref{tab:asve}, this modification results in a significant performance degradation, with a 4.32 dB drop in PSNR\textsubscript{mask} and a 0.0425 increase in LPIPS\textsubscript{mask}, underscoring the critical role of encoding visualizations in achieving high-fidelity results.

\noindent\textbf{Importance of Encoding 2D Visualization.} 
The visual encoding module of PoseFuse3D integrates 2D DWPose visualizations with rendered SMPL-X images. The 2D DWPose visualizations emphasize skeletal keypoint layouts, contributing to robust pose understanding. To assess its importance, we exclude the encoding of 2D visualizations, denoting this variant as `Non-2D'. This leads to a drop of 0.65 dB in PSNR\textsubscript{bbox} and a 13\% increase in LPIPS\textsubscript{bbox}, demonstrating the significance of encoding 2D visualizations in the visual encoding module.

\begin{table}[t]
\caption{\textbf{Ablation on the Visual Encoding.}}
\label{tab:asve}
\centering
\setlength\tabcolsep{3pt}
\resizebox{0.8\textwidth}{!}{
\begin{tabular}{lcccccc}
\toprule
\multirow{2}{*}{Model Variant}  & \multicolumn{6}{c}{Evaluation Metrics} \\ \cmidrule(l){2-7}
& PSNR$\uparrow$  & PSNR\textsubscript{bbox}$\uparrow$ & PSNR\textsubscript{mask}$\uparrow$  & LPIPS$\downarrow$  & LPIPS\textsubscript{bbox}$\downarrow$  & LPIPS\textsubscript{mask}$\downarrow$\\ \midrule

Non-Vis & 19.63 & 16.02 & 14.69  & 0.2097  & 0.1232 & 0.0889 \\
Non-2D & 21.71 & 18.65 & 17.28  & 0.1438  & 0.0738 & 0.0531 \\
Full & \textbf{22.14} & \textbf{19.30} & \textbf{18.01}  & \textbf{0.1330}  & \textbf{0.0653} & \textbf{0.0464} \\
\bottomrule
\end{tabular}
}
\vspace{-3pt}
\end{table}
\subsection{Ablation on SMPL-X Encoder}
\label{sec:ase}
\begin{table}[t]
\caption{\textbf{Ablation on SMPL-X Encoder.}}
\label{tab:abse}
\centering
\setlength\tabcolsep{3pt}
\resizebox{0.8\textwidth}{!}{
\begin{tabular}{lcccccc}
\toprule
\multirow{2}{*}{Model Variant}  & \multicolumn{6}{c}{Evaluation Metrics} \\ \cmidrule(l){2-7}
& PSNR$\uparrow$  & PSNR\textsubscript{bbox}$\uparrow$ & PSNR\textsubscript{mask}$\uparrow$  & LPIPS$\downarrow$  & LPIPS\textsubscript{bbox}$\downarrow$  & LPIPS\textsubscript{mask}$\downarrow$\\ \midrule

Non-JA & 22.07 & 19.24 & 17.95  & 0.1348  & 0.0659 & 0.0466 \\
Non-VA & \textbf{22.15} & 19.27 & 17.99  & 0.1374  & 0.0667 & 0.0470 \\
Full & 22.14 & \textbf{19.30} & \textbf{18.01}  & \textbf{0.1330}  & \textbf{0.0653} & \textbf{0.0464} \\
\bottomrule
\end{tabular}
}
\end{table}
We conduct an additional ablation study on the SMPL-X encoder to justify our design.

\noindent\textbf{Joint Aggregation.}
The SMPL-X encoder extracts joint motion and position features from 3D space and projects them onto the 2D image plane via an attention mechanism, providing spatial human body motion cues. To assess the impact of this design, we remove the joint aggregation module, denoted as `Non-JA'. As shown in Table~\ref{tab:abse}, this leads to a 0.06 dB drop in both PSNR\textsubscript{bbox} and PSNR\textsubscript{mask}, emphasizing the importance of 3D joint aggregation for accurate body control representation.

\noindent\textbf{Vertex Aggregation.}
We also apply a separate attention mechanism to aggregate vertex information into the 2D image plane. To examine its necessity, we remove the vertex attention module. We denote this variant as `Non-VA'. As reported in Table~\ref{tab:abse}, this leads to a noticeable degradation in performance across all LPIPS scores, including a 0.0040 rise in LPIPS and a 0.0014 increase in LPIPS\textsubscript{bbox}. These results demonstrate the significance of incorporating 3D vertex information for effective control representation from SMPL-X.

\subsection{Supplementary Visualizations}
\label{sec:av}
\begin{figure*}[t]
  \vspace{-5pt}
  \centering
  \includegraphics[width=1.0\linewidth]{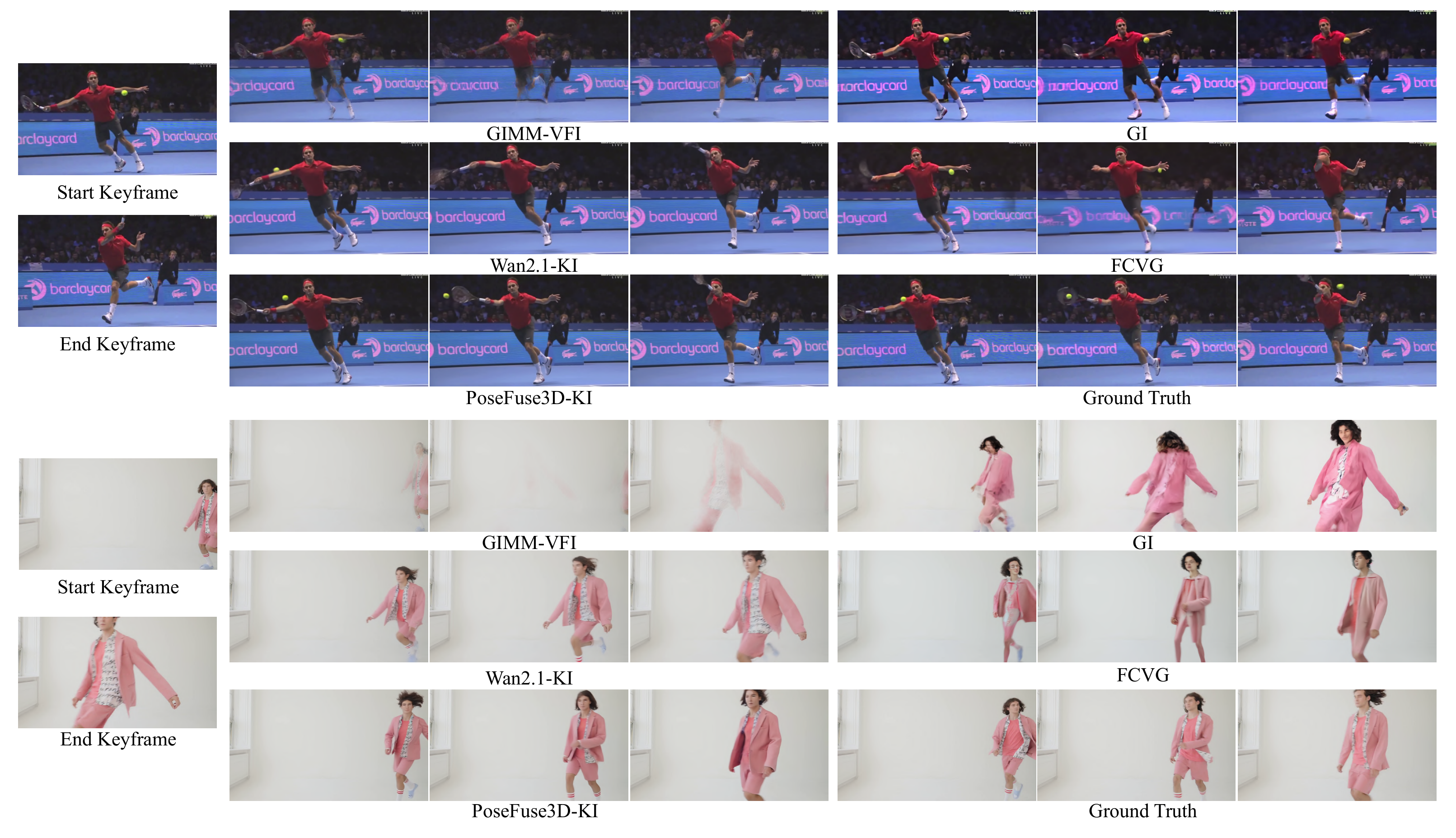}
  \caption{\textbf{Additional Qualitative Results on CHKI-Video.} }
  \label{fig:supp_add}
  \vspace{-8pt}
\end{figure*}

\noindent\textbf{Additional Qualitative Comparisons.}
We present additional qualitative comparisons with other interpolation methods in Figure~\ref{fig:supp_add}. Our PoseFuse3D-KI framework consistently produces more plausible interpolations, closely capturing real-world dynamics observed in the ground truth.

\begin{figure*}[t]
  \vspace{-5pt}
  \centering
  \includegraphics[width=1.0\linewidth]{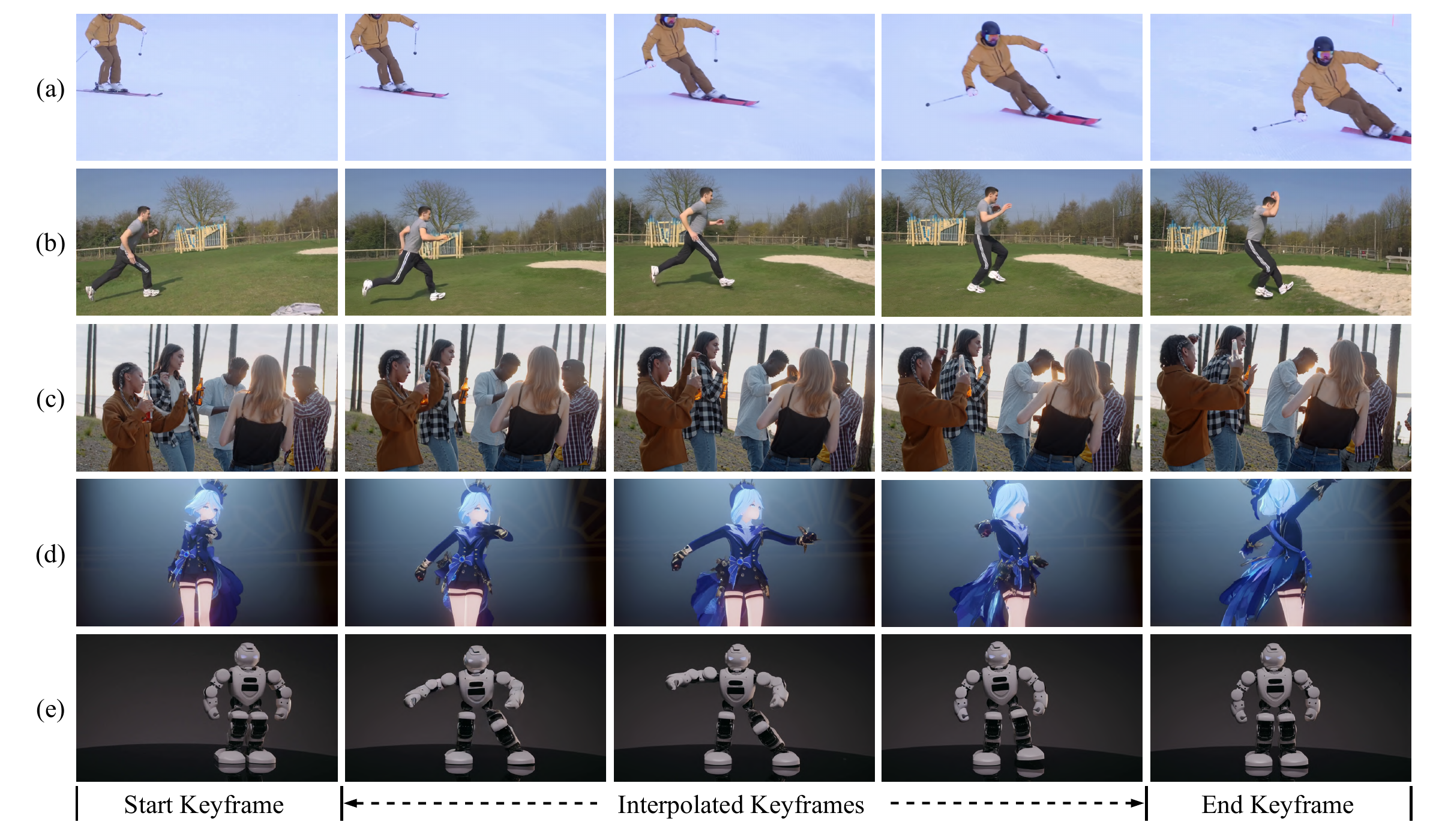}
  \caption{\textbf{Qualitative Results on Corner Cases.} (a) Person off-center. (b) Camera motion. (c) Multiple persons. (d) Virtual animation. (e) Robotics video.}
  \label{fig:corner_case}
  \vspace{-8pt}
\end{figure*}

\noindent\textbf{Robustness across Corner Cases.}
PoseFuse3D-KI exhibits robust interpolation performance across diverse corner cases. As shown in Figure~\ref{fig:corner_case}, we evaluate scenarios such as off-center subjects, camera motion, multiple persons, virtual animation, and robotics videos. Our method consistently produces controllable interpolations with plausible and temporally coherent motion.

\subsection{Runtime Comparison}
\label{sec:runtime}
\begin{table}[t]
\caption{\textbf{Runtime Comparison.} 
}
\small
\label{tab:runtime}
\centering
\setlength\tabcolsep{15pt}
\resizebox{0.8\textwidth}{!}{
\begin{tabular}{lccc}
\toprule
% \hline
Methods & GI~\cite{wang2024generative} & FCVG~\cite{zhu2024generative}  & PoseFuse3D-KI (Ours)  \\ \midrule
Runtime (s) & 975 & 523 & \textbf{212} \\
\bottomrule
% \hline 			
\end{tabular}
}
\end{table}
To assess computational efficiency, we compare the runtime of PoseFuse3D-KI with existing diffusion-based baselines. The comparison measures the time required to interpolate 25 frames at a resolution of $1024\times576$, using the same GPU equipped with 80 GB memory. As summarized in Table~\ref{tab:runtime}, PoseFuse3D-KI exhibits the shortest runtime, confirming its efficiency advantage over competing methods.

\subsection{Limitations}
\label{sec:lim}
There are several known limitations to our method. First, PoseFuse3D-KI relies on accurate SMPL-X estimations to generate reliable 3D control signals. Therefore, it inherits the limitations of the 3D human model estimators, where inaccurate predictions can degrade the quality of interpolated results. Additionally, our method, while offering strong control via 3D and 2D fusion, still depends on the base diffusion model’s generative priors. As a result, output quality is influenced by the model’s learned behavior and inherits its high GPU memory demands. Finally, our method does not explicitly model human-object interactions, which may lead to artifacts or misaligned object motion in scenarios involving close interaction with external objects.

\subsection{Broader Impacts}
\label{sec:bi}
Our proposed method, PoseFuse3D-KI, enables accurate and controllable human-centric keyframe interpolation, with applications in areas such as human animation and video generation. By integrating explicit 3D information from human models and 2D pose cues, our framework supports 3D-informed and semantically meaningful guidance for interpolating realistic human motion across frames. This technique not only enriches creative workflows but also opens new opportunities for research in human motion understanding and video synthesis. While powerful, our method shares common limitations of generative models and may pose risks if misused to produce manipulated or deceptive human videos, highlighting the importance of responsible use and ethical safeguards.

% \clearpage

\end{document}